\documentclass[]{spie}  

\usepackage{threeparttable}
\usepackage{amsmath,amsfonts,amssymb}
\usepackage{subcaption}
\usepackage{graphicx}
\usepackage{cite} 
\usepackage{epsfig}
\usepackage{nccmath}
\usepackage{colortbl}
\usepackage{tabularx}
\usepackage{relsize}
\usepackage{pifont}
\usepackage{booktabs} 
\usepackage{multirow}
\usepackage{multicol}
\usepackage{adjustbox}
\usepackage{float}
\usepackage{makecell}
\usepackage{tabu}
\usepackage[colorlinks=true, allcolors=blue]{hyperref}
\usepackage[capitalize]{cleveref}
\usepackage{algorithm}

\newcolumntype{P}[1]{>{\centering\arraybackslash}p{#1}}

\usepackage[table]{xcolor}
\usepackage{siunitx}
\usepackage{array}
\definecolor{retrieval}{HTML}{FB5368} 
\definecolor{percohort}{HTML}{46B1E1} 
\definecolor{baselinegray}{gray}{0.4} 

\newcommand{\aucgray}[1]{\cellcolor{gray!10}#1}
\newcommand{\aucblue}[1]{\cellcolor{percohort!20}#1}
\newcommand{\aucpink}[1]{\cellcolor{retrieval!20}#1}

\newcommand{\textpink}[1]{\textcolor{retrieval}{#1}}
\newcommand{\textblue}[1]{\textcolor{percohort}{#1}}
\newcommand{\textgray}[1]{\textcolor{baselinegray}{#1}}

\title{Cohort-Aware Agents for Individualized Lung Cancer Risk Prediction Using a Retrieval-Augmented Model Selection Framework}

\author[a]{Chongyu Qu}
\author[b]{Allen J. Luna}
\author[a,b]{Thomas Z. Li}
\author[a]{Junchao Zhu}
\author[a]{Junlin Guo}
\author[a]{Juming Xiong}
\author[b]{Kim L. Sandler}
\author[a,b]{Bennett A. Landman}
\author[a]{Yuankai Huo}

\affil[a]{Vanderbilt University, Nashville TN 37235, USA}
\affil[b]{Vanderbilt University Medical Center, Nashville TN 37232, USA}

\authorinfo{Further author information: (Send correspondence to Yuankai Huo)\\
E-mail: yuankai.huo@vanderbilt.edu}

\begin{document}
\maketitle

\begin{abstract}
Accurate lung cancer risk prediction remains challenging due to substantial variability across patient populations and clinical settings—\textbf{no single model performs best for all cohorts}. To address this, we propose personalized lung cancer risk prediction agent that dynamically selects the most appropriate model for each patient by combining cohort-specific knowledge with modern retrieval and reasoning techniques. Given a patient’s CT scan and structured metadata—including demographic, clinical, and nodule-level features—the agent \textbf{first} performs cohort retrieval using FAISS-based similarity search across nine diverse real-world cohorts to identify the most relevant patient population from a multi-institutional database. \textbf{Second}, a Large Language Model (LLM) is prompted with the retrieved cohort and its associated performance metrics to recommend the optimal prediction algorithm from a pool of eight representative models, including classical linear risk models (e.g., Mayo, Brock), temporally-aware models (e.g., TD-VIT, DLSTM), and multi-modal computer vision based approaches (e.g., Liao, Sybil, DLS, DLI). This two-stage agent pipeline—retrieval via FAISS and reasoning via LLM—enables dynamic, cohort-aware risk prediction personalized to each patient’s profile. Building on this architecture, the agent supports flexible and cohort-driven model selection across diverse clinical populations, offering a practical path toward individualized risk assessment in real-world lung cancer screening.
\end{abstract}


\section{Introduction}
\label{sec:intro}
Accurate risk prediction for lung cancer remains a critical challenge in medical AI, particularly in large-scale screening~\cite{rivera2013establishing,lokhandwala2017costs} programs and incidental nodule evaluation~\cite{lokhandwala2017costs,gould2015recent,mazzone2022evaluating,macmahon2017guidelines,detterbeck2013executive}. Despite the wide variety of available predictive models—ranging from classical statistical risk scores~\cite{mcwilliams2013probability,swensen1997probability} to sophisticated computer vision based approaches~\cite{li2023time,gao2020time,gao2021deep,gao2021cancer}—recent studies have consistently shown that no single model achieves optimal performance across all clinical contexts or patient populations~\cite{li2025performance}. Cohort-specific characteristics, variability in imaging protocols~\cite{li2018ct,choe2019deep,qu2023abdomenatlas}, and differences in patient demographics routinely lead to substantial distribution shifts~\cite{ktena2024generative,li2024abdomenatlas,kull2017beyond}. Consequently, predictive models often experience large performance degradation when applied out-of-domain (OOD)~\cite{lasko2024probabilistic,elsahar2019annotate,zhu2025asign,zhu2025magnet,zhu2025cross,zhu2023anti,youssef2023external,hu2023image,li2021performance}.

Retrieval-augmented modeling~\cite{blattmann2022retrieval,gao2023retrieval} approaches have recently emerged as a promising paradigm to address such distributional challenges. By retrieving relevant contextual information from external knowledge bases or historical databases, these methods provide critical context to inform downstream reasoning tasks. In domains such as natural language processing, retrieval-augmented generation (RAG)\cite{lewis2020retrieval} techniques have demonstrated substantial improvements in task-specific accuracy and generalization~\cite{liu2023learning,zakka2024almanac}. In clinical prediction, where understanding a patient's specific cohort context can be decisive, retrieval strategies offer similar potential by leveraging patient-specific metadata and imaging features to identify clinically relevant reference cohorts, enabling more informed downstream model selection decisions.

\begin{figure}[!h]
    \centering
    \includegraphics[width=1\textwidth]{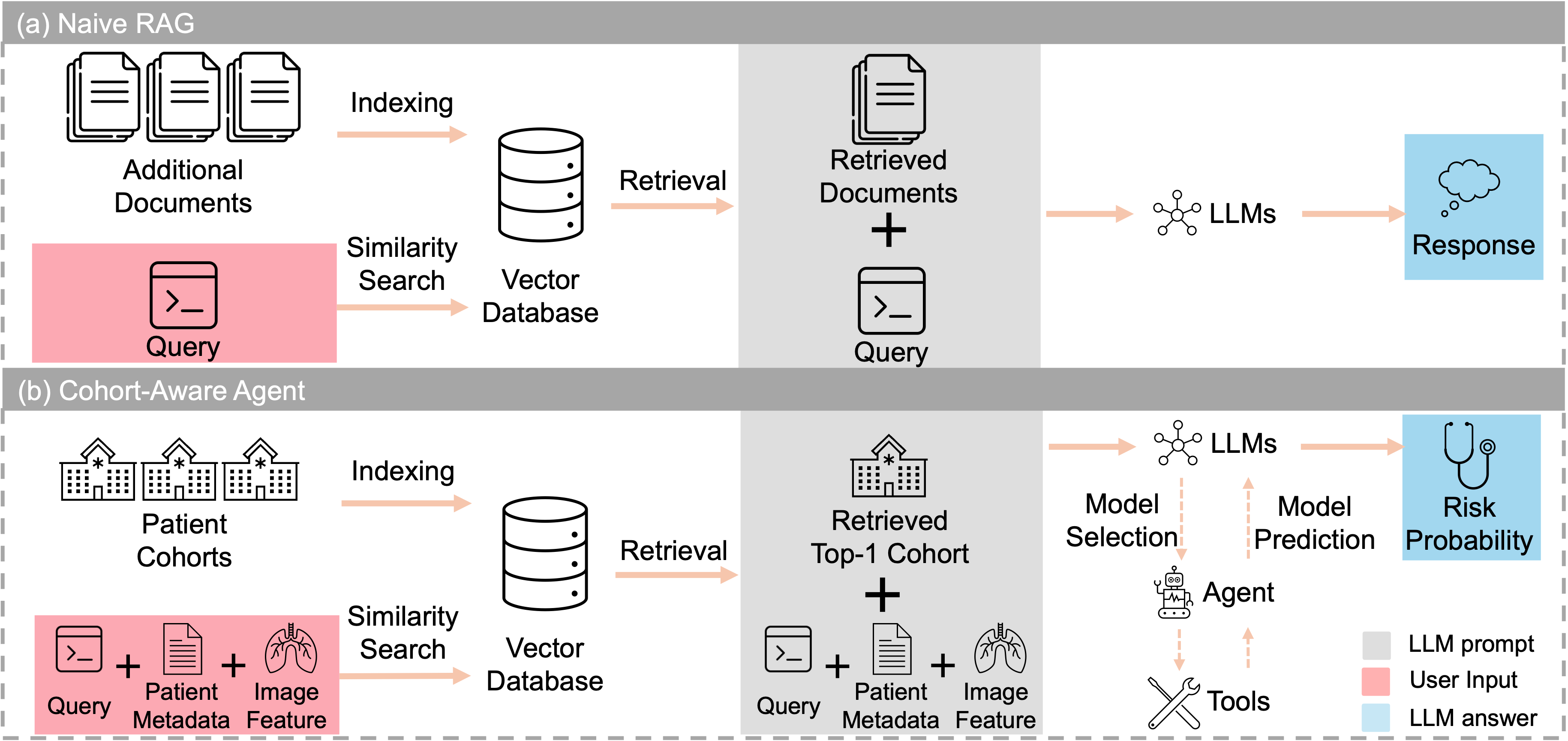}
    \caption{\textbf{Cohort‑aware agent vs. naive RAG.} (a) Naive RAG retrieves context primarily based on a user query and applies a fixed reasoning process. In contrast, our (b) cohort-aware agent uses patient metadata and imaging features to retrieve the most relevant reference cohort. The agent then constructs an LLM prompt that includes the full patient profile, the retrieved cohort, and the prediction task. The LLM selects an appropriate model by invoking tools and returns a personalized lung cancer risk prediction.}
    \label{fig:rag}
\end{figure}

However, to fully realize this potential, flexible reasoning systems that dynamically leverage retrieved context are necessary. Agent-based AI systems~\cite{durante2024agent,huang2022language,peng2023check} particularly those powered LLMs~\cite{bubeck2023sparks} have recently demonstrated remarkable flexibility in orchestrating multimodal reasoning~\cite{peng2023check} and decision making tasks~\cite{gong2023mindagent}. Despite these advances, existing agent frameworks remain predominantly designed for general-purpose tasks. Their suitability for clinical prediction remains limited, especially in complex scenarios such as lung cancer risk assessment~\cite{petit2019lung,guida2018assessment}, where substantial variations across patient cohorts severely constrain the effectiveness of general-purpose reasoning strategies. In clinical contexts characterized by heterogeneous patient populations and substantial distribution shifts, relying on a single unified reasoning or predictive approach is inadequate. This highlights a key limitation of current agent-based frameworks: the absence of mechanisms to adapt decision-making to cohort-specific clinical contexts in a personalized and data-driven way.

To address this gap, we propose Cohort-Aware Agents, a retrieval-augmented framework for individualized lung cancer risk prediction. Rather than applying a fixed model across all patients, our approach automatically selects the most appropriate prediction algorithm by combining cohort retrieval with LLM-based reasoning. The agent follows a two-stage architecture inspired by clinical decision-making. In the \textbf{first} stage, it retrieves the most relevant reference cohort by comparing a patient’s CT scan and structured clinical metadata—including demographic~\cite{tammemagi2013selection}, clinical, and nodule-level features—with a large, multi-institutional database using FAISS-based~\cite{douze2024faiss} similarity search. This allows the agent to match each patient to a representative clinical subpopulation, capturing subtle patterns in data distribution. In the \textbf{second} stage, a LLM is prompted with the retrieved cohort’s characteristics and associated model performance summaries to identify the most suitable prediction algorithm.  
Rather than applying a fixed model, our agent dynamically selects from eight risk models, including classical statistical scores (e.g., Mayo~\cite{mcwilliams2013probability}, Brock~\cite{swensen1997probability}), temporally-aware models (e.g., TD-VIT~\cite{li2023time}, DLSTM~\cite{gao2020time}), and computer vision based approaches (e.g., Liao~\cite{liao2019evaluate}, Sybil~\cite{mikhael2023sybil}, DLS~\cite{gao2021cancer}, DLI~\cite{gao2021deep}). Figure \ref{fig:rag} illustrates this overall framework by contrasting our cohort-aware agent with a naive RAG baseline. Our cohort-aware agent enables personalized, context-aware model selection that adapts to cohort-specific characteristics, providing a practical and generalizable solution for individualized lung cancer risk assessment. In summary, our agent makes the following contributions:

$\bullet$ \textit{Cohort-aware patient retrieval:} We develop a FAISS-based retrieval strategy that identifies the most relevant reference cohort for each patient by leveraging structured metadata and CT imaging features, enabling population-contextualized reasoning.

$\bullet$ \textit{Retrieval-augmented model selection:} We design a two-stage agent pipeline where an LLM receives a patient’s profile and their most similar cohort retrieved from a FAISS-indexed database, and selects the optimal risk prediction model by referencing precomputed cohort-specific performance metrics.

$\bullet$\textit{Personalized risk prediction across diverse cohorts:}We demonstrate the effectiveness of our agent on nine real-world lung cancer screening cohorts and eight representative models, showing improved generalizability and individualized performance through cohort-driven model selection.

\section{Methods}
\label{sec:method}
\begin{figure}[t]
    \centering
    \includegraphics[width=1\textwidth]{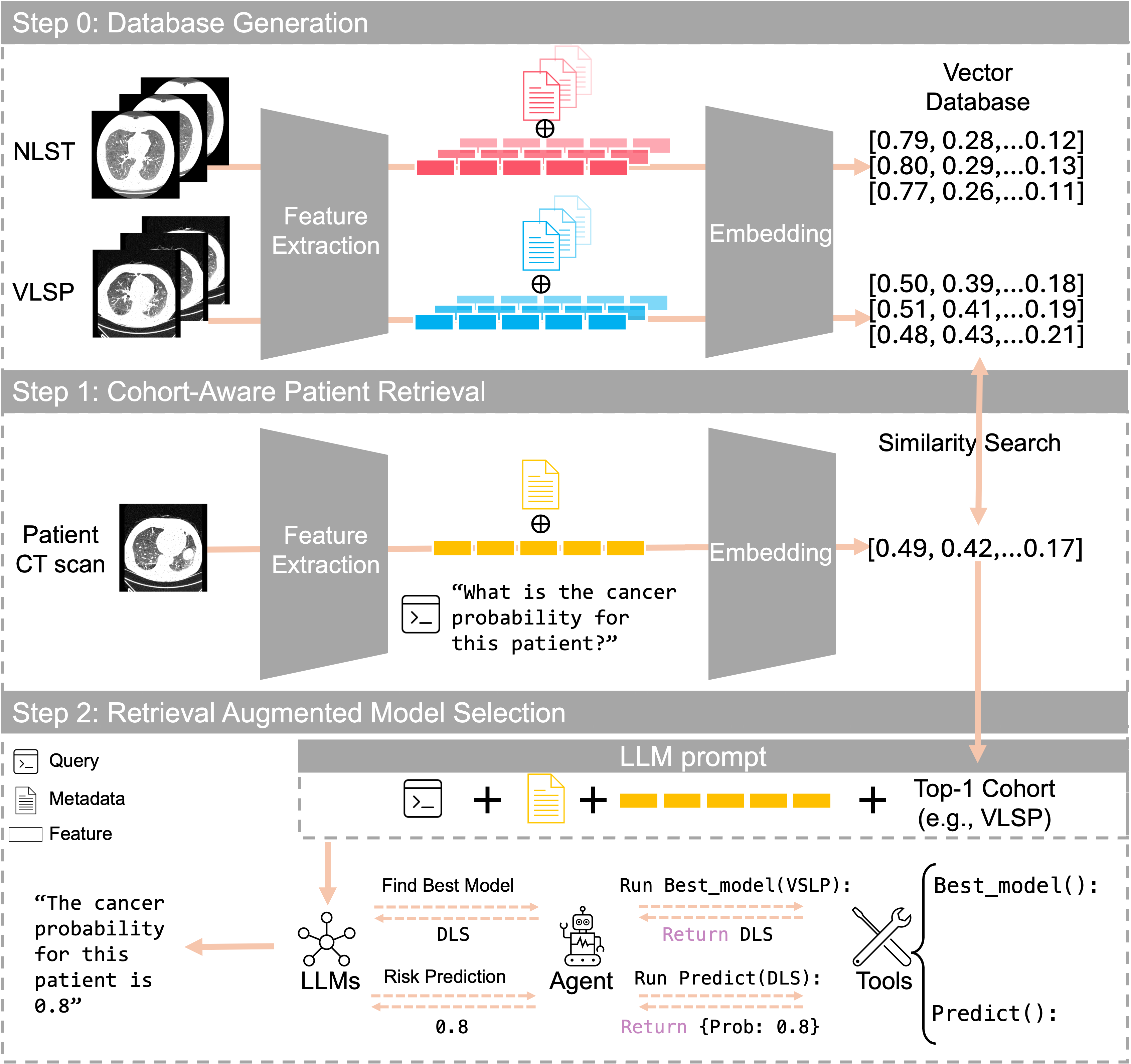}
    \caption{\textbf{The proposed Cohort-Aware Agents framework.} Patient metadata and imaging features are combined to retrieve the top-1 cohort. This cohort, together with the task query, guides the agent to select the optimal model via tool use and generate an individualized lung cancer risk prediction.}
    \label{fig:framework}
\end{figure}

\textbf{Overview.} We propose a Cohort-Aware Agent for individualized lung cancer risk prediction, built on a two-stage retrieval-augmented model selection framework, illustrate in Figure~\ref{fig:framework}. Prior to inference, CT volumes from multiple cohorts are processed with the DeepLungScreening (DLS) pipeline to extract imaging features, which are concatenated with structured metadata and embedded into a FAISS-indexed vector database (Step~0). For a new patient, the same pipeline generates a query vector that is compared with the database to retrieve the top-1 cohort. During \textbf{cohort-aware patient retrieval} (\S\ref{sec:cohort_retrieval}, Step~1), this cohort provides population context for downstream reasoning. The patient’s metadata, CT-derived features, the top-1 cohort, and the task query together form the LLM prompt. During \textbf{retrieval-augmented model selection} (\S\ref{sec:model_selection}, Step~2), the LLM issues structured requests to the agent. The agent invokes tools to select the optimal model for that cohort, runs inference, and returns an individualized lung cancer risk probability. This design enables personalized, cohort-aware predictions and supports flexible, cohort-driven model selection across diverse clinical populations.

\subsection{Cohort-aware patient retrieval}
\label{sec:cohort_retrieval}

To provide population-contextualized reasoning, the agent first determines the most relevant reference cohort for each incoming patient. This step addresses the observation that predictive performance can vary markedly across patient populations due to distribution shifts in imaging protocols, demographic composition, and clinical characteristics. By retrieving a clinically similar cohort, the agent grounds its subsequent model selection in patterns that are specific to that population.

Each patient is represented by a multi-dimensional feature vector $\mathbf{x} \in \mathbb{R}^d$, defined as the concatenation of structured metadata (e.g., age, gender, smoking history) and imaging-derived features extracted from the CT scan. A similarity search module compares this patient vector against a pre-constructed vector database of prior cases, each associated with a known cohort label.

Let $\mathcal{D} = \{(\mathbf{x}_i, c_i)\}_{i=1}^N$ denote the indexed database, where $\mathbf{x}_i \in \mathbb{R}^d$ is the feature representation of the $i$-th patient, and $c_i \in \{1, \dots, C\}$ is the corresponding cohort label.

To identify similar cases, the agent performs approximate nearest-neighbor search over $\mathcal{D}$ using FAISS. The top-$k$ retrieved neighbors of $\mathbf{x}$ are defined as:
\begin{equation}
\mathcal{N}_k(\mathbf{x}) = \text{TopK}_i \left( -\left\lVert \mathbf{x} - \mathbf{x}_i \right\rVert_2 \right),
\end{equation}
where $\left\lVert \cdot \right\rVert_2$ denotes the Euclidean (L2) norm. Cosine similarity may also be used depending on feature normalization.

The final retrieved cohort label $\hat{c}$ is determined by majority vote over the cohort labels of the retrieved neighbors:
\begin{equation}
\hat{c} = \text{mode} \left( \{ c_i \mid (\mathbf{x}_i, c_i) \in \mathcal{N}_k(\mathbf{x}) \} \right).
\end{equation}

\subsection{Retrieval-augmented model selection}
\label{sec:model_selection}

Once the top-1 cohort $\hat{c}$ has been identified, the agent proceeds to select the most appropriate lung cancer risk prediction model based on cohort context and patient-specific information. This stage leverages the reasoning capabilities of a LLM to coordinate model selection via tool use.

The LLM is prompted with the user-issued task query, the patient’s metadata and imaging-derived features (concatenated as $\mathbf{x}$), and the retrieved cohort $\hat{c}$. It returns a model selection decision as follows:
\begin{equation}
m^* = \text{LLM}(\text{query}, \mathbf{x}, \hat{c}),
\end{equation}
where $m^*$ denotes the identifier of the selected model. The agent then invokes this model and applies it to the patient’s feature vector to generate a risk prediction:
\begin{equation}
\hat{y} = m^*(\mathbf{x}),
\end{equation}
where $\hat{y}$ is the predicted lung cancer risk probability. This completes the second stage of the framework, enabling cohort-adaptive, individualized inference.

\section{Data, Model and Experimental Setting}
\subsection{Cohort Overview}
To ensure consistent image quality across cohorts, we applied a standardized set of preprocessing criteria. Specifically, we excluded CT scans with severe imaging artifacts, slice thickness $\geq$5,mm, or incomplete lung fields in the field of view~\cite{gao2021technical}. All scans were de-identified using the MIRC Anonymizer~\cite{rsna_mirc_dicom_anonymizer} to remove protected health information. These quality and privacy controls help ensure comparability across institutions and enable robust representation learning from imaging data.

Our reference database comprises nine real-world lung cancer cohorts collected across institutions and clinical contexts, each with structured metadata, chest CT imaging, and diagnostic labels. The \textit{VLSP} cohort comes from the Vanderbilt Lung Screening Program and represents a low-dose CT screening population. The \textit{LI-VUMC}~\cite{li2025curating} cohort is a longitudinal imaging and biomarker cohort at VUMC, covering both high-risk and screening-detected cases. From the Multi-Center Lung (MCL) project, we incorporate four institution-specific cohorts: \textit{MCL-VUMC} (Vanderbilt University Medical Center), \textit{MCL-UPMC} (University of Pittsburgh Medical Center), \textit{MCL-UCD} (University of Colorado Denver), and \textit{MCL-DECAMP}~\cite{billatos2019detection} (DECAMP consortium), representing prospectively enrolled incidental nodule populations across sites. The \textit{BRONCH} cohort includes patients undergoing diagnostic bronchoscopy with pathology-adjudicated outcomes. We also include two subsets from the National Lung Screening Trial (NLST)~\cite{national2011reduced}: \textit{NLST-test}~\cite{ardila2019end}, a held-out split aligned with the original trial’s test set, and \textit{NLST-test-nodule}, a subset of \textit{NLST-test} with at least one nodule meeting the NLST positive definition.

All cohorts provide structured metadata suitable for similarity-based cohort retrieval (e.g., age, gender, BMI, smoking status, scan date). When nodule-level descriptors are unavailable, we rely on image-level features and cohort-level metadata; outcome labels follow each dataset’s ground-truth protocol (pathology, clinical coding, or longitudinal follow-up). This consistency enables robust patient representations and supports cohort-aware retrieval and model selection within our framework.

\subsection{Model Pool}

Our agent selects from a pool of eight lung cancer risk prediction models spanning diverse modeling paradigms and clinical assumptions. We group them into three categories:

\textbf{Linear risk models.} The Mayo~\cite{swensen1997probability} and Brock~\cite{mcwilliams2013probability} models are logistic-regression scores widely used in practice, leveraging variables such as nodule size, age, smoking status, and spiculation. They offer interpretability and solid baselines but are constrained by hand-crafted feature sets~\cite{revel2004two,oxnard2011variability,devaraj2017use} and the absence of learned imaging representations.

\textbf{Temporally-aware models.} TD-ViT~\cite{li2023time} and DLSTM~\cite{gao2020time} incorporate longitudinal CT to capture nodule evolution across time. TD-ViT encodes continuous time/scan position into a transformer, while DLSTM uses recurrent gating to model temporal dynamics. These models are applicable only when multiple timepoints are available~\cite{rudin2021small}.

\textbf{Computer vision-based models.} Liao~\cite{liao2019evaluate}, Sybil~\cite{mikhael2023sybil}, DeepLungScreening (DLS)~\cite{gao2021cancer}, and DeepLungIPN (DLI)~\cite{gao2021deep} learn high-dimensional predictors from CT, optionally fused with structured metadata. Sybil operates directly at the scan level without manual nodule annotations; DLS targets screening populations and integrates CT with clinical data elements; DLI targets indeterminate/incidental nodules with CT plus clinical variables~\cite{kammer2019compensated}. Because input requirements differ, not all models are applied to all cohorts.

Model selection is performed adaptively by the agent, conditioned on available inputs and historical performance on retrieved cohorts.

\subsection{Implementation details.} We include 3,750 patients from nine cohorts, with per-cohort sizes ranging from 104 to 868. A random 30\% of patients are held out for validation, while the remaining 70\% form the retrieval database. During evaluation, each validation patient retrieves from this database for cohort-aware model selection. Patient representations are constructed by concatenating structured metadata (e.g., age, gender, BMI, smoking status) with pooled imaging features. The imaging features are extracted using the DLS pipeline, which produces a $5 \times 128$ feature map per case. This map is pooled and re-weighted by a factor of 0.1 before fusion with metadata. Similarity search is performed using FAISS with cosine distance, retrieving the top-$k = 15$ neighbors.
Model selection is performed using TinyLlama-1.1B-Chat-v1.0 (FP16)~\cite{zhang2024tinyllama}, prompted with (i) the retrieved top-1 cohort, (ii) patient's metadata and imaging-derived features, and (iii) user-issued task query. The LLM returns the name of the most suitable model, which is then used to produce the final risk prediction. The full agent pipeline—including retrieval, prompting, and inference—runs on a single NVIDIA A6000 GPU (48GB), without the need for distributed computation.
\section{Results}
\label{sec:result}
\subsection{Cohort-Aware Retrieval Performance}
To evaluate the retrieval component of our agent, we assess its ability to correctly identify the ground-truth cohort for each validation patient based on top-1 neighbor vote. As described in Section~\ref{sec:model_selection}, all experiments use a fixed top-$k = 15$ for FAISS retrieval and apply a feature weight of 0.1 when combining metadata and imaging features.

The imaging features are extracted using the DLS pipeline, which outputs a 5$\times$128 feature map per patient. We experiment with two aggregation strategies: flattening the feature map into a single 640-dimensional vector, or applying average pooling across the temporal axis to produce a 128-dimensional embedding. These representations are then concatenated with structured metadata to form the final retrieval vector. We also compare L2 and cosine similarity metrics.

\begin{table}[h]
  \centering
  \caption{Top-1 accuracy of cohort retrieval under different input configurations and similarity metrics.}
  \label{tab:retrieval}
  \small
  \setlength{\tabcolsep}{25pt}
  \resizebox{\linewidth}{!}{%
  \begin{tabular}{l c c c}
    \toprule
    Input Type & Feature Aggregation & Similarity Metric & Top-1 Accuracy \\
    \midrule
    Metadata only            & \textemdash{} & L2     & 0.639 \\
    Metadata + Feature       & Flattened     & L2     & 0.222 \\
    Metadata + Feature       & Pooled        & L2     & 0.595 \\
    Metadata + Feature       & Pooled        & Cosine & \textbf{0.667}\\
    \bottomrule
  \end{tabular}}
\end{table}

The results are summarized in Table~\ref{tab:retrieval}.  Metadata alone provides a strong baseline with a top-1 accuracy of 0.639. Adding flattened imaging features significantly reduces accuracy to 0.222, likely due to high-dimensional noise and scale mismatches that negatively affect L2-based similarity. Aggregating the imaging features via average pooling improves stability and brings the accuracy back up to 0.595, though still slightly below the metadata-only baseline. The best performance is achieved when pooled imaging features are combined with cosine similarity, which normalizes vector magnitudes and better balances the influence of metadata and imaging inputs. This configuration yields a top-1 retrieval accuracy of 0.667, surpassing all other settings. These results highlight the importance of both appropriate feature aggregation and distance metric selection when integrating heterogeneous inputs for cohort retrieval.

\begin{table}[h]
  \centering
  \caption{\textbf{Top-1 Cohort Retrieval Confusion Matrix.} 
  Results are from the \emph{Metadata + Pooled Feature + Cosine} configuration. 
  Rows denote ground-truth cohorts, and columns show the retrieved top-1 cohorts. 
  Diagonal entries correspond to correctly retrieved cases, with an overall accuracy of 0.667.}
  \label{tab:retrieval_confusion}
  \small
  \setlength{\tabcolsep}{5pt}
  \resizebox{\linewidth}{!}{%
  \begin{tabular}{lcccccccccc}
    \toprule
    & BRONCH & MCL\_VUMC & MCL\_UPMC & MCL\_DECAMP & MCL\_UCD & VLSP & LI-VUMC & NLST\_test\_nodule & NLST\_test & correct / total (acc) \\
    \midrule
    BRONCH             & \cellcolor{gray!10}109 & 0 & 0 & 0 & 0 & 0 & 0 & 0 & 0 & 109 / 109 (1.000) \\
    MCL\_VUMC          & 0 & \cellcolor{gray!10}63 & 1 & 0 & 18 & 0 & 0 & 0 & 0 & 63 / 82 (0.768) \\
    MCL\_UPMC          & 0 & 22 & \cellcolor{gray!10}3 & 0 & 6 & 0 & 0 & 0 & 0 & 3 / 31 (0.097) \\
    MCL\_DECAMP        & 0 & 14 & 0 & \cellcolor{gray!10}19 & 3 & 0 & 0 & 0 & 0 & 19 / 36 (0.528) \\
    MCL\_UCD           & 0 & 19 & 0 & 0 & \cellcolor{gray!10}13 & 0 & 0 & 0 & 0 & 13 / 32 (0.406) \\
    VLSP               & 0 & 0 & 0 & 0 & 0 & \cellcolor{gray!10}256 & 0 & 0 & 1 & 256 / 257 (0.996) \\
    LI-VUMC            & 1 & 0 & 0 & 0 & 0 & 0 & \cellcolor{gray!10}45 & 0 & 15 & 45 / 61 (0.738) \\
    NLST\_test\_nodule & 0 & 0 & 0 & 0 & 0 & 0 & 2 & \cellcolor{gray!10}48 & 205 & 48 / 255 (0.188) \\
    NLST\_test         & 0 & 0 & 0 & 0 & 0 & 0 & 2 & 65 & \cellcolor{gray!10}193 & 193 / 260 (0.742) \\
    \midrule
    Overall            & \multicolumn{9}{c}{} & 749 / 1123 (0.667) \\
    \bottomrule
  \end{tabular}}
\end{table}
To better understand retrieval behavior across different cohorts, we present a confusion matrix in Table~\ref{tab:retrieval_confusion}, computed using the best-performing configuration (Metadata + Pooled Feature + Cosine). In this matrix, each row corresponds to the ground-truth cohort of a validation patient, while each column shows the top-1 cohort retrieved by the similarity searching. Diagonal entries indicate correct retrieval.

Perfect retrieval is observed for cohorts such as BRONCH and VLSP, which are highly distinct in both metadata and imaging space. In contrast, cohorts like MCL\_UPMC and NLST\_test\_nodule are more frequently misassigned, often confused with adjacent cohorts that share similar population characteristics. These patterns suggest that retrieval quality is influenced by inter-cohort similarity in the embedding space, which in turn affects downstream model selection accuracy.

\subsection{Model Selection Performance}
\begin{figure}[!h]
    \centering
    \includegraphics[width=0.6\textwidth]{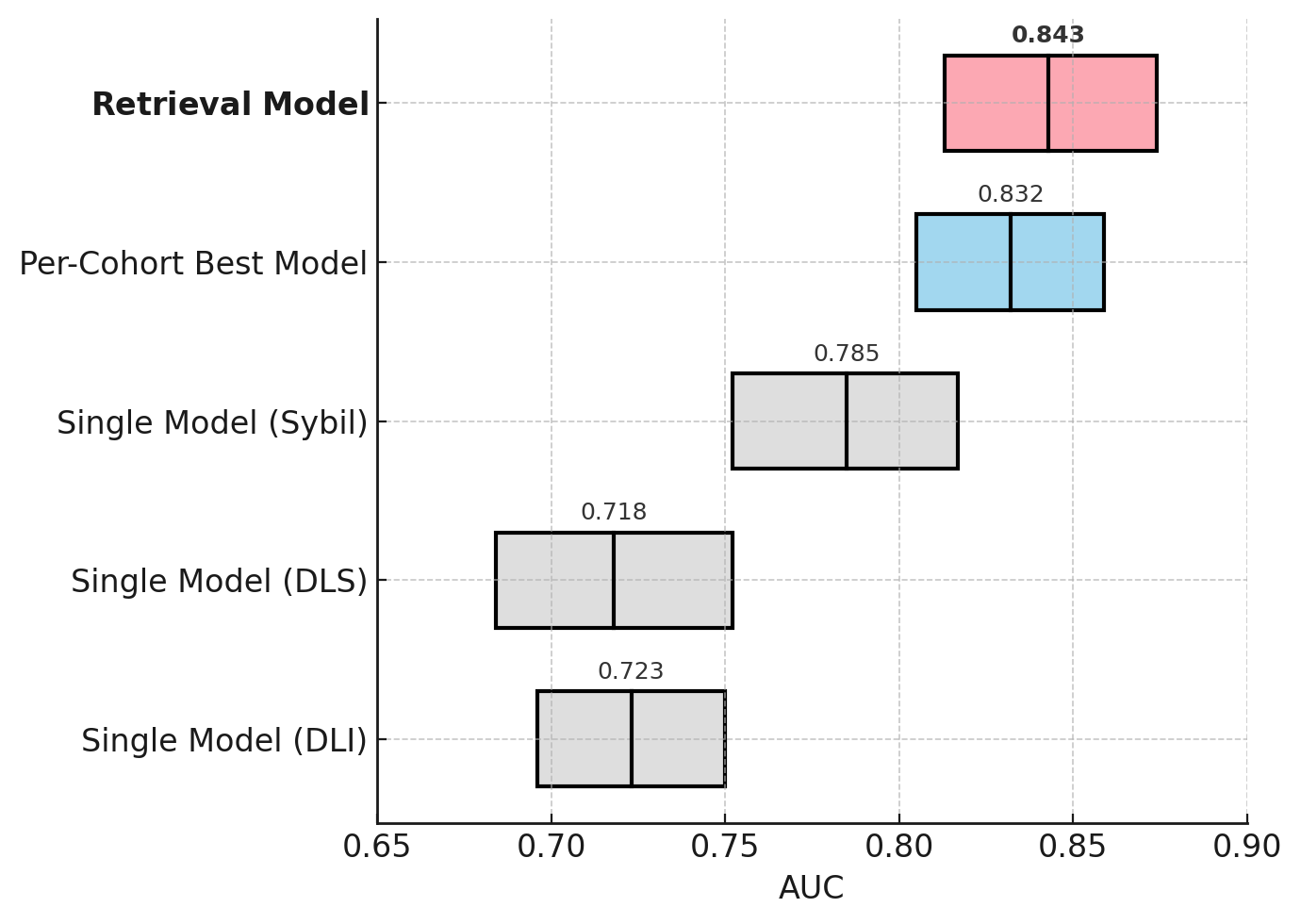}
    \caption{Overall AUC (Middle line) and 97.5 \% CI (Box) under Different Model Selection Strategies.}
    \label{fig:auc}
\end{figure}

\begin{table}[h]
  \centering
  \caption{\textbf{Cohort-wise AUC and Inference Time (s) under Different Model Selection Strategies.} The first three columns correspond to single-model baselines, \textgray{Single Model (DLI, DLS, Sybil)}, where a fixed model is used across all cohorts. The \textblue{Per-Cohort Best Model} selects the best-performing model for each cohort based on pre-computed validation performance. The \textpink{Retrieval Model} reports the performance of our cohort-aware agent, which retrieves a top-1 similar cohort for each validation case and applies that cohort’s best-performing model}
  \label{tab:cohort_results}
  \small
  \setlength{\tabcolsep}{5pt}
  \resizebox{\linewidth}{!}{%
  \begin{tabular}{l*{5}{cc}}
    \toprule
    \multirow{2}{*}{Cohort} &
    \multicolumn{2}{c}{Single Model (DLI)} &
    \multicolumn{2}{c}{Single Model (DLS)} &
    \multicolumn{2}{c}{Single Model (Sybil)} &
    \multicolumn{2}{c}{Per-Cohort Best Model} &
    \multicolumn{2}{c}{Retrieval Model} \\
    \cmidrule(lr){2-3}\cmidrule(lr){4-5}\cmidrule(lr){6-7}\cmidrule(lr){8-9}\cmidrule(lr){10-11}
     & AUC & Time (s) & AUC & Time (s) & AUC & Time (s) & AUC & Time (s) & AUC & Time (s) \\
    \midrule
    BRONCH             & \aucgray{0.609} & 0.55 & \aucgray{0.643} & 0.55 & \aucgray{0.657} & 1355.51 & \aucblue{0.657} & 1296.04 & \aucpink{0.660} & 1331.15 \\
    MCL\_VUMC          & \aucgray{0.827} & 0.39 & \aucgray{0.765} & 0.39 & \aucgray{0.829} & 875.76  & \aucblue{0.829} & 827.01  & \aucpink{0.804} & 1064.71 \\
    MCL\_UPMC          & \aucgray{0.983} & 0.33 & \aucgray{0.880} & 0.33 & \aucgray{0.923} & 199.02  & \aucblue{0.983} & 0.50    & \aucpink{1.000} & 0.30    \\
    MCL\_DECAMP        & \aucgray{0.738} & 0.34 & \aucgray{0.753} & 0.34 & \aucgray{0.654} & 365.96  & \aucblue{0.753} & 0.35    & \aucpink{0.761} & 0.32    \\
    MCL\_UCD           & \aucgray{0.938} & 0.33 & \aucgray{0.805} & 0.34 & \aucgray{0.801} & 349.33  & \aucblue{0.938} & 0.34    & \aucpink{0.892} & 0.46    \\
    VLSP               & \aucgray{0.510} & 0.63 & \aucgray{0.811} & 0.63 & \aucgray{0.783} & 3325.91 & \aucblue{0.811} & 0.67    & \aucpink{0.810} & 0.62    \\
    LI-VUMC            & \aucgray{0.824} & 0.37 & \aucgray{0.545} & 0.37 & \aucgray{0.725} & 1011.29 & \aucblue{0.824} & 0.38    & \aucpink{0.842} & 0.36    \\
    NLST\_test\_nodule & \aucgray{0.545} & 0.61 & \aucgray{0.627} & 0.61 & \aucgray{0.853} & 1166.94 & \aucblue{0.853} & 1173.90 & \aucpink{0.982} & 498.98  \\
    NLST\_test         & \aucgray{0.534} & 0.64 & \aucgray{0.634} & 0.64 & \aucgray{0.838} & 1156.02 & \aucblue{0.838} & 1197.63 & \aucpink{0.840} & 2144.22 \\
    \midrule
    Overall            & \aucgray{0.723} & 4.49 & \aucgray{0.718} & 4.50 & \aucgray{0.785} & 10805.84 & \aucblue{0.832} & 4,496.82 & \aucpink{0.843} & 5041.12 \\
    \bottomrule
  \end{tabular}}%

\end{table}
We evaluate the effectiveness of different model selection strategies in Table~\ref{tab:cohort_results} and Figure~\ref{fig:auc} report the AUC,inference time and 97.5\% confidence interval (CI) across cohorts under five configurations. 

The first three strategies apply a \textit{single model—DLI, DLS, or Sybil}—uniformly to all patients, regardless of cohort identity. Among these baselines, Sybil achieves the highest overall AUC (0.785), followed by DLI (0.723) and DLS (0.718). However, Sybil also incurs a substantially higher total inference time (10,805.84 seconds), reflecting the computational cost of its more complex architecture. In contrast, DLI and DLS are significantly more efficient, with total runtime under 5 seconds per cohort.

The Per-Cohort Best Model assigns to each cohort the model that historically achieved the best validation performance. This strategy relies on cohort identity and applies the corresponding best-performing model to all patients within that cohort. It improves the overall AUC to 0.832 while maintaining low total runtime (4,496.82 seconds),  since a subset of cohort-optimal models are lightweight architectures such as DLI or DLS, though several cohorts still favor more complex models like Sybil. This result demonstrates the benefit of population-specific model allocation when accurate cohort information is available.

Our proposed \textit{Retrieval Model} delivers performance comparable to the \textit{Per\mbox{-}Cohort Best Model}. Across cohorts, the mean $\Delta\mathrm{AUC}$ (Retrieval $-$ Per\mbox{-}Cohort Best) is $0.0117$ with a 95\% bootstrap confidence interval $[-0.0136,\,0.0457]$ based on 1{,}000 resamples. For each validation case, the model retrieves the most similar cohort from the vector database via similarity search and then applies that cohort’s historically best model. This selection introduces minimal overhead and yields a competitive runtime (5041.12 seconds), comparable to the \textit{Per\mbox{-}Cohort Best Model}. These results support individualized, cohort\mbox{-}aware model selection in heterogeneous patient populations.

Figure~\ref{fig:auc} further summarizes the overall AUC and 97.5\% confidence intervals under each strategy, confirming that the retrieval-based approach outperforms uniform and delivers performance comparable to cohort-specific baselines.

\section{Conclusion}
We propose a cohort-aware agent for individualized lung cancer risk prediction, which leverages population-contextualized reasoning to select optimal predictive models for each patient. By retrieving clinically similar cohorts via a shared patient representation and applying a retrieval-based model selection strategy, our method accounts for distributional heterogeneity across institutions, imaging protocols, and demographics. Extensive evaluation across nine cohorts shows that our agent consistently achieves state-of-the-art performance, outperforming all single-model baselines and delivering results comparable to cohort-specific oracle assignments. This framework offers a scalable and adaptable solution for personalized risk prediction in real-world deployment settings, and illustrates the potential of retrieval-augmented agents to support data-driven generalization across diverse clinical populations.

\acknowledgements
\begin{flushleft}
This research was supported by the National Institutes of Health (NIH) through grants F30CA275020, 2U01CA152662, R01CA253923 (Landman \& Maldonado), R01CA275015 (Maldonado \& Lenburg), U01CA152662 (Grogan), U01CA196405 (Maldonado), and P30CA068485-29S1, as well as the National Science Foundation (NSF) through CAREER 1452485 and grant 2040462. Additional support was provided by the Vanderbilt Institute for Surgery and Engineering through T32EB021937-07, the Vanderbilt Institute for Clinical and Translational Research via UL1TR002243-06, the Pierre Massion Directorship in Pulmonary Medicine, and the American College of Radiology Fund for Collaborative Research in Imaging (FCRI) Grant. This manuscript was polished using AI-assisted editing (ChatGPT) with a “rephrase” prompt; no scientific content was generated or altered.  
\end{flushleft}
\bibliographystyle{spiebib} 
\bibliography{report.bib}

\end{document}